\documentclass[conference]{IEEEtran}
\IEEEoverridecommandlockouts
\usepackage{cite}
\usepackage{amsmath,amssymb,amsfonts}
\usepackage{algorithmic}
\usepackage{graphicx}
\usepackage{textcomp}
\usepackage{xcolor}
\usepackage{algorithm}
\usepackage{algorithmic}
\usepackage{subcaption}
\usepackage{url}
\def\BibTeX{{\rm B\kern-.05em{\sc i\kern-.025em b}\kern-.08em
    T\kern-.1667em\lower.7ex\hbox{E}\kern-.125emX}}
\begin{document}

\newcommand{\fm}[1]{\textcolor{purple}{FM: #1}}
\newcommand{\hs}[1]{\textcolor{blue}{HS: #1}}

\title{DELATOR: Money Laundering Detection via Multi-Task Learning on Large Transaction Graphs \\
}

\author{\IEEEauthorblockN{Henrique S.\ Assumpção}
\IEEEauthorblockA{
\textit{Universidade Federal de Minas Gerais} \\
henriquesoares@dcc.ufmg.br}
\and
\IEEEauthorblockN{Fabrício Souza}
\IEEEauthorblockA{
\textit{Universidade Federal de Minas Gerais} \\
fabricio.souza@dcc.ufmg.br}
\and
\IEEEauthorblockN{ Leandro Lacerda Campos}
\IEEEauthorblockA{
\textit{InterMind (Inter Bank)}\\
\textit{Universidade Federal de Minas Gerais} \\
leandro.campos@bancointer.com.br}
\and
\IEEEauthorblockN{Vinícius T.\ de Castro Pires}
\IEEEauthorblockA{
\textit{InterMind (Inter Bank)}\\
\textit{Universidade Federal de Minas Gerais} \\
vinicius.pires@bancointer.com.br}
\and
\IEEEauthorblockN{Paulo M.\ Laurentys de Almeida}
\IEEEauthorblockA{
\textit{InterMind (Inter Bank)}\\
paulo.laurentys@bancointer.com.br}
\and
\IEEEauthorblockN{Fabricio Murai}
\IEEEauthorblockA{
\textit{Worcester Polytechnic Institute}\\
\textit{Universidade Federal de Minas Gerais}\\
fmurai@wpi.edu}
}

\maketitle

\begin{abstract}
Money laundering has become one of the most relevant criminal activities in modern societies, as it causes massive financial losses for governments, banks and other institutions. Detecting such activities is among the top priorities when it comes to financial analysis, but current approaches are often costly and labor intensive partly due to the sheer amount of data to be analyzed. Hence, there is a growing need for automatic anti-money laundering systems to assist experts. In this work, we propose DELATOR, a novel framework for detecting money laundering activities based on graph neural networks that learn from large-scale temporal graphs. DELATOR provides an effective and efficient method for learning from heavily imbalanced graph data, by adapting concepts from the GraphSMOTE framework and incorporating elements of multi-task learning to obtain rich node embeddings for node classification. DELATOR outperforms all considered baselines, including an off-the-shelf solution from Amazon AWS by 23\% with respect to AUC-ROC. We also conducted real experiments that led to the discovery of 7 new suspicious cases among the 50 analyzed ones, which have been reported to the authorities.
\end{abstract}


\section{Introduction}\label{Section:Introduction}
Money laundering is a general term referring to a myriad of different criminal activities that seek to legitimize illicit financial gains or to conceal the origin of money transferred to illegal entities, e.g., terror financing and drug trafficking~\cite{FATF}. Although varied in nature, money laundering activities can be categorized into three well-defined stages that represent the full process of legitimizing a series of illegal economic activities: (i) placement, (ii) layering, and (iii) integration~\cite{Ebikake}. Placement refers to the initial process of introducing the illicitly gained assets into the legitimate financial system, by removing obvious traces of illegality.
Layering is one of the most important and complex stages of money laundering. It consists of a series of transactions with no purpose other than to conceal the illicit origin of the money. Finally, integration aims to integrate these assets into the legal economy. 

Most western countries implement a rule-based regulation that all banks must follow in order to comply with legislation \cite{BACEN,FATF}. Such rules vary in nature, but can often describe numerical thresholds for flagging suspicious transactions, e.g., receiving or sending an amount that surpasses $n$ times the client's income. Traditionally, flagged clients are manually analyzed by an anti-money laundering (AML) team of experts, who try to determine the existence of extraordinary circumstances that could justify the threshold violation (e.g., a client sold a property they owned). Based on that analysis and the current legislation, the team decides whether to report the client to the authorities or not, however, this manual inspection of flagged clients is an expensive and time-consuming process. Not surprisingly, in Brazil, banks have 45 days to file reports regarding the transactions performed in a given month. 

Such rule-based AML systems have three main limitations. First, it is hard to prioritize flagged clients, except by ranking them according to the number of rule violations\footnote{Although a violation severity score could be computed for each rule, it is not obvious how to combine them into a single score.}. A good prioritization scheme could significantly expedite the reporting of criminals and could provide side information to help experts close cases unlikely to be reported. Second, there is potentially a large number of clients ``just below'' thresholds for being flagged that will never be analyzed. Despite not violating any given rule, altogether, the set of transactions made by a client could provide evidence of a crime. Potential losses for a false negative are immense since the financial institution could face significant legal backlash. Last, 
the fact that some common transaction patterns (e.g., Smurfing \cite{Smurfs}) occur in money laundering activities underlines the limitations of rules that apply to clients individually, namely, they do not account for structural information in the network formed by all transactions.

Therefore, there is a notable demand for new data-driven tools for money laundering detection that escape the rule-based approaches. In this paper, we will focus on building a robust data-driven framework for detecting money laundering activity in the layering stage by adapting different machine learning models and novel approaches for graph-structured data. We propose a multi-task learning framework named DELATOR for detecting money laundering in dynamic financial transaction networks. For each graph snapshot, it generates client (node) embeddings based both on the unsupervised link prediction task and on a supervised edge regression task where the labels are the transaction values. Next, the embeddings of each client are concatenated over the snapshots. DELATOR then generates synthetic suspicious clients in the embedding space to oversample the minority class. Last, it trains a supervised classifier for predicting the probability of a given client being involved in money laundering. 

Thus, we summarize our contributions as follows:
\begin{itemize}
    \item We propose a scalable and effective AML framework that experimentally outperforms state-of-the-art methods and has a relatively simple implementation.
    \item DELATOR is among the first methods for detecting money laundering activity on large transaction graphs that are temporal, heterogeneous, and have extremely imbalanced target classes. By leveraging different aspects of the network, we are able to simplify the data modeling and create a method that performs well on a real-world large dataset.
    \item We evaluate our framework in a real setting by performing data-driven inference on millions of bank accounts, which ultimately led to the detection of multiple suspicious clients, that were then reported to the authorities for potential involvement in money laundering activities. To the best of our knowledge, this is the first system of its kind to be employed successfully in the context of Brazilian banks.

\end{itemize}
\section{Related Work} \label{Section: Related Work}

\subsection{Graph Neural Networks} \label{SubSection: Graph Neural Networks}
Our work is mostly related to graph representation learning, and more specifically, learning node representations on a latent space based on unsupervised or semi-supervised tasks. In recent years, models based on Graph Neural Networks (GNNs) \cite{DBLP:journals/corr/KipfW16} have taken a prominent role in the context of learning latent representations for graph-structured data. GraphSAGE \cite{DBLP:journals/corr/HamiltonYL17} was among the first unsupervised learning frameworks that could yield an inductive model for graphs. Its loss function is based on random walks, and it encourages nearby nodes to have similar embeddings, while ensuring that distant nodes have dissimilar embeddings.
The Graph Attention architecture \cite{gat} has also been successful in many different graph-related tasks, and it utilizes masked self-attention during the message passing mechanism, allowing the network to learn which neighbors to prioritize when aggregating information.

There are also many recent techniques specifically designed for temporal graphs. EvolveGCN \cite{evolvegcn} combines a Recurrent Neural Network (RNN) with a GNN to account for the temporal aspect of dynamic graphs. The authors propose two alternative approaches for these networks: (i) consider the GNN weight matrix as the hidden state of the dynamical system, or (ii) use the GNN weight matrix as input of the dynamical system. Both methods allow for a time-aware representation on the embedding space.

The aforementioned approaches are designed for homogeneous graphs, but there also are many attempts to generalize GNNs for heterogeneous graphs. The authors who proposed RGCN \cite{rgcn} were among the first to extend the message passing mechanism of the original GNN to heterogeneous graphs, by learning on subgraphs independently created for each relation and then aggregating the information across all relations. More recent approaches like HGT \cite{DBLP:journals/corr/abs-2003-01332} combine elements of the Transformer architecture \cite{DBLP:journals/corr/VaswaniSPUJGKP17} with GNNs in order to model heterogeneous topological relations that can also evolve through time. These approaches generally focus on training the models directly on a downstream task. One exception is the DGL-KE \cite{DGL-KE} framework implemented on the Deep Graph Library \cite{dgl}, which provides a general framework for learning from knowledge graphs based on a generalization of the unsupervised link prediction task that yields both node and relation embeddings.

\subsection{Class Imbalance} \label{Subsection: Class Imbalance}
Class imbalance refers to a significant difference in the number of instances of each class and is a characteristic displayed by many real-world datasets. It tends to bias model predictions towards the majority classes. For this reason., class imbalance has long been a relevant research topic within the machine learning community \cite{class_imbalance,Japkowicz02theclass}.

There are many strategies for dealing with this problem, some of which seek to adjust class size through over- or undersampling, i.e., making the majority classes smaller and the minority classes bigger. SMOTE \cite{DBLP:journals/corr/abs-1106-1813} is one of the most popular oversampling algorithms, and it addresses the problem by interpolating samples belonging to minority classes with their nearest neighbors in order to create new synthetic samples. However, SMOTE does not consider topological information into consideration, and it is dependent on the distance metric chosen to determine the nearest neighbors. GraphSMOTE \cite{DBLP:journals/corr/abs-2103-08826} tries to solve this potential issue by adapting the SMOTE algorithm to better suit graph representation learning applications. GraphSMOTE first extracts node embeddings from the graph by using a single-layer GNN, and then applies SMOTE to each minority class in order to generate synthetic nodes to balance the previously underrepresented classes. GraphSMOTE also creates an edge generator in order to connect these nodes to the network, by training a GNN-based classifier that aims to be good at reconstructing the original adjacency matrix of the network. The framework then proceeds to train a final classifier for node classification on the downstream task.

Other approaches like Pick-and-Choose \cite{liu2021pickandchoose} try to deal with the imbalance problem by creating synthetic edges between nodes in the minority class, however, the authors of \cite{brasnam21} show that this technique has a significantly lower performance on highly imbalanced datasets. After an extensive analysis of the available literature, we decided to adapt some of the ideas from GraphSMOTE -- which was originally designed for static graphs -- to our framework in order to improve the process of learning information about our dynamic network.
\subsection{Money Laundering Detection} \label{Subsection: Money Laundering Detection}
Detecting money laundering activity on financial transaction networks can be seen as a particular case of fraud detection on graph-based data. In this more general context, most of the problems are related to imbalanced scenarios, where a small minority of individuals are actually involved in the target illicit activity. 

The authors of \cite{smurfs_detection} provide a simple and efficient method for detecting suspicious activity specifically related to money laundering, by employing a series of standard database join operations to detect sub-graphs that follow the Smurf patterns. The authors of \cite{Alarab2022} provide a GNN-based approach for money laundering detection on graphs, by employing an LSTM network coupled with a graph convolutional network in order to simultaneously model the topological and temporal relationships between nodes on the graph. Finally, the authors of \cite{DBLP:journals/corr/abs-1908-02591} provide a comparative analysis between classical machine learning methods -- such as Random Forests and Logistic Regression -- and GNN approaches for detecting money laundering activity in a large bitcoin dataset, and find that the GNN models with skip connections have the best performance amongst the considered baselines. However, all of the aforementioned models do not directly deal with the imbalance problem that is prominent in our data.

\section{Data Modeling} \label{Section: Data Modeling}
In this section, we explain the model we consider in this work and discuss a potential, but more complex alternative. 

The available transaction data consists of a multiset $\mathcal{E}$ of monetary transactions with partition $\mathcal{E} = \bigcup_{t=1}^\mathcal{T}\mathcal{E}^t$, where each $\mathcal{E}^t$ represents all transactions made at timestep $t \in \{1,2,...,\mathcal{T}\}$, a set $\mathcal{V}$ of users and a set $\mathcal{X} = \{(\textbf{x}_1,y_1),...,(\textbf{x}_n,y_n)\}$, where each $\textbf{x}_i \in \mathbb{R}^d$ represents a given user's attributes. Each $y_i$ represents the target variable indicating if the user was suspected of being involved in money laundering activity. Each element $e = ((v,u),w,c) \in \mathcal{E}^t$ represents a transaction from user $v$ to user $u$, of amount $w$ and type $c \in \{1,2,...,m\}$, executed at timestep $t$, and there is no restriction on the amount of transactions between users. Since the data is dynamic in nature, it is natural to model it as a \textit{dynamic graph}, more specifically as a \textit{Discrete-time dynamic graph} (DTDG), which is a sequence of static graph snapshots taken at certain time intervals. In our context, each snapshot contains the same set of nodes and node attributes. This data structure allows for two alternative modeling approaches, which will be referred to as \textbf{A1} and \textbf{A2}.

 \textbf{A1}: We can model the data as a \textbf{homogeneous directed weighted} DTDG, i.e., a sequence $S_{\text{homo}} = \{G^1,G^2,...,G^\mathcal{T}\}$ of graphs, where each $G^t \in S_{\text{homo}}$ represents the network at time $t$. In order to obtain a simple homogeneous graph at each snapshot $t$, we create a new edge set $E^{t} = \{((v,u),w')|w' = \sum_{e \in \mathcal{E}^t:e_1=(v,u)} w\}$ -- where $e_i$ represents the $i$-th element of the tuple $e$ --, that aggregates all edges between each pair of nodes $(v,u)$ by setting $w'$ as the total amount transacted from $v$ to $u$. We can then define each snapshot as $G^t = (\mathcal{V},E^{t},\mathcal{X})$.
 This modeling aggregates transactions with different types into a single transaction, thus simplifying the graph structure when compared to approach \textbf{A2}.

\textbf{A2}: We can model the data as a \textbf{heterogeneous directed weighted} DTDG, i.e., a sequence $S_{\text{hetero}} = \{G^1,G^2,...,G^\mathcal{T}\}$ of graphs, where each $G^t \in S_{\text{hetero}}$ represents the network at time $t$. In this scenario, each snapshot is called a \textit{knowledge graph}, and it consists of a set of relations that represent each transaction type. More formally, we can define each snapshot as $G^t = \{G_1^t,G_2^t,...,G_m^t\}$, where each $G_c^t = (\mathcal{V},\mathcal{E}_c^t,\mathcal{X})$ is the relational graph for snapshot $t$ and transaction type $c$, and $\mathcal{E}_c^t = \{e \in \mathcal{E}^t|e_3=c\}$.
In contrast to \textbf{A1}, approach \textbf{A2} provides a more complete description of the network, however, this comes at the cost of being more complex and costly to model computationally. 
 \section{Proposed Framework} \label{Section: Proposed Framework}
In this section, we provide a brief overview of DELATOR, and then proceed to detail each step of the framework. DELATOR follows approach \textbf{A1} for modeling the data, and it is comprised of three main steps:
\begin{enumerate}
    \item This step consists of a multi-task learning algorithm to obtain node embeddings for each graph snapshot. We first train a GNN model that aims to optimize the \textit{link prediction loss}, i.e., a loss function that tends to enforce embedding similarity between connected nodes. Next, we train a second GNN model that aims to optimize the \textit{edge regression loss}, i.e., a task for predicting the edge weights. The second network also generates node embeddings, that are then concatenated with the embeddings generated by the first GNN and passed through a \textit{Multi-Layered Perceptron} (MLP) to obtain the final prediction for the edge weight.
    These networks are trained individually on their respective tasks and each generates node embeddings for all graph snapshots. We concatenate such embeddings into a single representation per snapshot, and then concatenate the representations across all snapshots in order to obtain a single time-aware description of the nodes in euclidean space;
    \item Next, we create synthetic nodes to oversample the minority class, by applying the SMOTE algorithm directly to the embedding space on the training set in order to obtain new embeddings for the minority class.
    \item We finally proceed to train a classifier on the aggregated oversampled data that predicts the probability of a user being suspected of being involved in money laundering.
\end{enumerate}

\begin{figure}[htbp]
\centerline{\includegraphics[width=90mm]{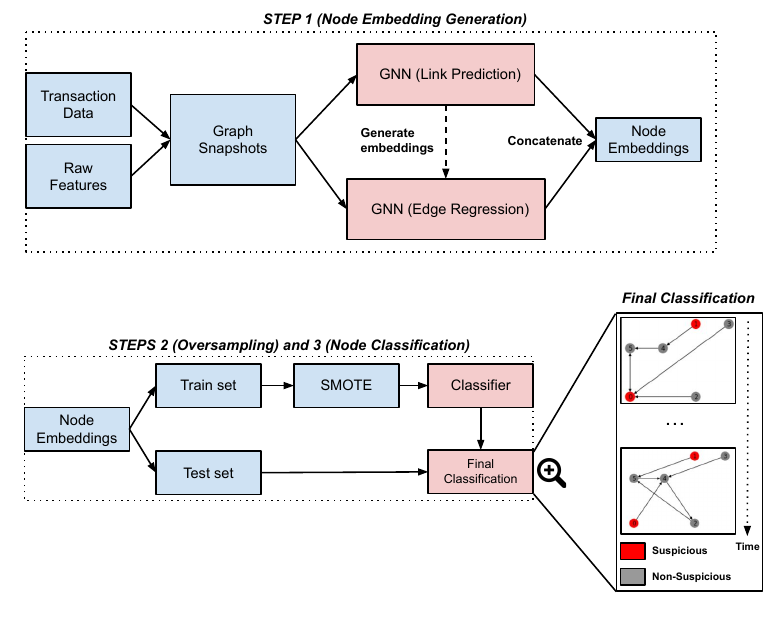}}
\caption{Overview of DELATOR.}
\label{fig:delator_overview}
\end{figure}

We highlight the existence of major challenges for modeling the money laundering detection task using current machine learning methods on graphs, both due to the inherently difficult nature of the problem at hand and also due to specific characteristics of our data, namely: (i) the transaction data is dynamic in nature, which significantly increases the complexity of the proposed task, (ii) there are multiple transaction types in the data, and thus the resulting graph is highly heterogeneous, which significantly increases the complexity of the problem since there are few methods in the literature suited for dynamic heterogeneous graphs, and even fewer that specifically target money laundering detection, (iii) the target classes are extremely imbalanced since only a small percentage of users are actually labeled as involved in money laundering activity, and (iv) the data is considerably large in size, thus any proposed method will need to be scalable, in addition to being effective.

Regarding the temporal aspect of the data (challenge (i)), we only have access to a relatively small time interval, and thus methods based on Recurrent Neural Networks have proven to be ineffective. For this reason, we chose to employ a simple concatenation procedure for the embeddings obtained at each snapshot that is able to consider the temporal aspect of our data. In order to address the other challenges, we propose a framework that has three main advantages: (i) by disregarding different edge types and modeling the data as a homogeneous simple graph, we significantly increase the framework's efficiency, since most modern approaches for heterogeneous graphs suffer from significant scaling issues when dealing with a high number of relations, and, as we will later discuss, this simplification does not interfere with the framework's efficacy on our dataset, (ii) we incorporate a strong oversampling technique to the framework, thus relieving the impact of class imbalance on the final classification task, and (iii) the framework is modular in nature, and thus it can be largely modified in order to adapt to different modeling scenarios. 

\subsection{Node Embedding Generation} \label{Subsection: Node Embedding Generation}
The first part of the framework consists of extracting node embeddings from the network. We adopt a multi-task learning approach and thus subdivide the training procedure into two steps.
\subsubsection{Link Prediction (LP)} \label{Subsubsection: Link Prediction}
First, we train a GNN network to obtain node embeddings based on an unsupervised loss for link prediction, i.e., predicting if two nodes are connected in the graph. Our framework allows for any GNN architecture to be used in this step, and we will present an example using the GraphSAGE architecture. Recall the sequence $S_{\text{homo}}$ defined in approach \textbf{A1}. For each graph snapshot $G^t \in S_{\text{homo}}$, the following equation describes the message passing mechanism for an arbitrary layer $l$ out of a total of $L$ layers of GraphSAGE: 
\begin{equation} \label{eq:sage_message_passing}
    h^{t,l}_v = f((W_l\cdot \text{AGG}(\{h^{t,l}_u, \forall u \in \mathcal{N}(v)\}))\mathbin\Vert (B_l\cdot h^{t,l-1}_v))
\end{equation}
In \eqref{eq:sage_message_passing}, $h^{t,l}_v$ denotes the embedding of node $v$ at layer $l$ and snapshot $t$, $W_{l}$ denotes a set of learnable weights, $B_{l}$ denotes a set of learnable biases, $\mathbin\Vert$ denotes the concatenation operator along the columns, AGG denotes an aggregation function, $\mathcal{N}(v)$ denotes the outgoing neighbours of node $v$ and $f$ denotes an activation function, e.g., ReLU, and $\cdot$ denotes matrix multiplication. We initialize $h^{t,0}_v = \textbf{x}_v$ for all nodes. For simplicity, we will refer to $h^{t,L}_v$ as $h^t_v$. We can then define the link prediction loss $\mathcal{L}^t_{\text{lp}}$ for a given snapshot $t$, that we wish to minimize, as follows:
\begin{align}
        L^t_{\text{pos}}(G^t) &=\sum_{((v,u),w)\in E^t} -\log(\sigma((h^t_v)^\top \cdot h^t_u)) \label{eq:lp_pos} \\
        L^t_{\text{neg}}(G^t) &=\sum_{((v,u),w) \in \tilde{E}^t} -\log(1-\sigma((h^t_v)^\top \cdot h^t_u))) \label{eq:lp_neg}\\ 
        \mathcal{L}^t_{\text{lp}} &= \frac{L^t_{\text{neg}}(G^t) + L^t_{\text{pos}}(G^t)}{|E^t|+|\widetilde{E}^t|} \label{eq:lp_loss}
\end{align}
In \eqref{eq:lp_pos}-\eqref{eq:lp_loss}, $\sigma$ is the sigmoid function, $L^t_{\text{pos}}(G^t)$ denotes the log-likelihood of the link prediction for the current snapshot and $L^t_{\text{neg}}(G^t)$ denotes the log-likelihood of the link prediction for a negative sample of the current snapshot, i.e., we define a set $\tilde{E}^t$ of randomly selected edges, and $^\top$ denotes the transpose operator. This loss function is a direct adaptation of the GraphSAGE unsupervised loss, and it encourages the model to create embeddings that are similar for connected nodes while enforcing a distinct representation for disconnected nodes. The GNN model is trained consecutively on $S_{\text{homo}}$, i.e., at each epoch, we compute the embeddings and scores according to the aforementioned equations, and then perform one optimization step at each snapshot. Empirical testing showed that this method is preferable instead of optimizing w.r.t.\ the mean of the losses throughout all snapshots.
\subsubsection{Edge Regression (ER)} \label{Subsubsection: Edge Regression}
After training a GNN model for link prediction, we now train a model to perform edge regression, i.e., predict the value of a given transaction between two users. The model consists of a separate GNN that generates embeddings $z_v^{t}$ for all nodes, and we again note that this step can be performed with any GNN architecture. For an arbitrary edge $e=((v,u),w) \in E^t$, we obtain the predicted edge weight $\hat{w}(e)$ as follows:
\begin{equation}\label{eq:er_pred}
    \hat{w}(e) = \text{MLP}(h_v^{t} \mathbin\Vert z_v^{t} \mathbin\Vert h_u^{t} \mathbin\Vert z_u^{t})
\end{equation}
In \eqref{eq:er_pred}, the predicted edge weight is obtained by concatenating the link prediction and edge regression embeddings of both $v$ and $u$, and then passing them through a MLP network. We can then define the edge regression loss $\mathcal{L}^t_{\text{er}}$ for a given snapshot $t$, that we wish to minimize, as follows:
\begin{equation} \label{eq:er_loss}
\begin{split}
    l^t_{\text{er}}(w,\hat{w}) &= 
     \begin{cases}
       \frac{0.5 \cdot (\hat{w} - w)^2}{\gamma}, &\quad\text{if $|\hat{w} - w| < \gamma$}  \\
       |\hat{w} - w| - 0.5 \cdot \gamma, &\quad\text{otherwise} \\
     \end{cases} \\
     \mathcal{L}^t_{\text{er}} &= \frac{1}{|E^t|} \cdot \sum_{e=((v,u),w) \in E^t} l^t_{\text{er}}(w,\hat{w}(e))
\end{split}
\end{equation}
In \eqref{eq:er_loss}, $\gamma$ denotes a threshold hyperparameter, and $l^t_{\text{er}}(w,\hat{w})$ denotes the partial loss for a given edge. Equation \eqref{eq:er_loss} represents the smooth L1 loss function of the edge regression task. We also train this model on $S_{\text{homo}}$ in a consecutive fashion. After training the models, we obtain a single time-aware embedding for a given node $v$, denoted by $\eta(v)$, as follows:
\begin{equation} \label{eq:final_embedding}
    \eta(v) = \overset{\mathcal{T}}{\underset{t=1}{\mathbin\Vert}}(h_v^{t} \mathbin\Vert z_v^{t})
\end{equation}
Equation \eqref{eq:final_embedding} describes the procedure of concatenating the link prediction and edge regression embeddings for a given node to obtain a single embedding for each snapshot, and then concatenating these embeddings along all snapshots. We can then build a single embedding matrix $\textbf{H} \in \mathbb{R}^{|\mathcal{V}| \times (\mathcal{T} \cdot (d_{\text{LP}}+d_{\text{ER}}))}$, where $d_{\text{LP}}$ ($d_{\text{ER}}$) denotes the embedding size for the link prediction (edge regression) task, by concatenating the final embedding $\eta$ for all nodes along the rows, i.e., each row of $\textbf{H}$ is the corresponding vector $\eta$ for that given node.
\subsection{Oversampling}
We now seek to generate synthetic nodes to oversample the minority class. In this work we decided to adopt the SMOTE algorithm, however our framework is compatible with any oversampling methods that can use the generated embeddings.

The intuition behind SMOTE is to perform interpolation on samples from the minority class with their nearest neighbors in the embedding space of the same class, in order to generate new embeddings that are similar to the ones found in this minority class. Given a node $v \in \mathcal{V}$ belonging to a certain minority class, e.g., individuals involved in money laundering, consider the final embedding $\eta(v)$. SMOTE generates a new sample $v'$ according to the following equation:
\begin{equation}  \label{eq:smote}
    \eta(v') = (1-\lambda) \cdot \eta(v) + \lambda \cdot \eta(\text{nn}(v))
\end{equation}
In \eqref{eq:smote}, $\text{nn}(v)$ denotes the nearest neighbor of $v$ from the same class, according to the euclidean distance in the embedding space, and $\lambda \sim U(0,1)$ denotes a uniform random variable. Since the new node is generated via interpolation of nodes from the same class, we can label the new synthetic node $v'$ as belonging to the same class as $v$, thus we can artificially oversample the minority class to overcome the imbalanced scenario. We highlight that this oversampling step is only performed on the training data. We will refer to the oversampled aggregated embeddings matrix as $\textbf{H}'$.

\subsection{Node Classification} \label{Subsection: Node Classification}
The last step of DELATOR consists of a supervised classification task for detecting users suspected of being involved in money laundering activity. After oversampling the minority class, we employ LightGBM \cite{NIPS2017_6449f44a}, a state-of-the-art gradient boosting algorithm for supervised learning. The LightGBM model takes $\textbf{H}'$ as input and outputs a probability classification $\boldsymbol{\hat{Y}}$ of the users, i.e., a real value between 0 and 1 indicating the probability of a given user being suspicious. We highlight that our framework supports any supervised classifier at this step. Algorithm \ref{alg:delator_framework} describes the entire process of training DELATOR and predicting users involved in money laundering.
\begin{algorithm}[htbp]
   \caption{DELATOR Framework}
   \label{alg:delator_framework}
\begin{algorithmic}
   \STATE {\bfseries Input:} $S_{\text{homo}} = \{G^1,...,G^\mathcal{T}\}$, Number of epochs $N$
   \STATE {\bfseries Output:} Predicted probabilities $\boldsymbol{\hat{Y}}$
   \STATE Randomly initialize the weights for the GNNs of step 1, i.e., the GNN for link prediction and the GNN for edge regression
   \STATE Create train-test-validation splits based on the edges of each graph snapshot for the aforementioned tasks
   \FOR{$n \in \{1,2,...,N\}$}
   \FOR{$G^t \in S_{\text{homo}}$}
   \STATE Obtain the node embeddings for the current snapshot via LP according to \eqref{eq:sage_message_passing}
   \STATE Compute $\mathcal{L}^t_{\text{lp}}$ according to \eqref{eq:lp_pos}-\eqref{eq:lp_loss}
   \STATE Perform one optimization step
   \ENDFOR 
   \ENDFOR
   \FOR{$n \in \{1,2,...,N\}$}
   \FOR{$G^t \in S_{\text{homo}}$}
   \STATE Obtain the node embeddings for the current snapshot via ER according to \eqref{eq:sage_message_passing} 
   \STATE Compute the predicted edge weights for all edges according to \eqref{eq:er_pred}
   \STATE Compute $\mathcal{L}^t_{\text{er}}$ according to \eqref{eq:er_loss}
   \STATE Perform one optimization step
   \ENDFOR 
   \ENDFOR
   \STATE Compute the final embeddings according to \eqref{eq:final_embedding}, and then concatenate them to obtain $\textbf{H}$
   \STATE Create train-test-validation splits for $\textbf{H}$
   \STATE Oversample the minority class according to \eqref{eq:smote} to the train split to obtain $\textbf{H}'$
   \STATE Train a supervised classifier with $\textbf{H}'$ to obtain the final prediction $\boldsymbol{\hat{Y}}$
\end{algorithmic}
\end{algorithm}

\section{Experiments} \label{Section: Experiments}
In this section, we detail the experiments conducted to evaluate the performance of DELATOR. Specifically, we are interested in the following questions:
\begin{enumerate}
    \item Can DELATOR's simpler approach (\textbf{A1}) perform well on complex, heterogeneous relational networks?
    \item How well does DELATOR perform when compared to other state-of-the-art techniques for node embeddings and node classification?
    \item How sensitive is DELATOR w.r.t.\ its hyperparameters?
    \item How important is each component of the framework to its performance?
    \item Can DELATOR help the AML team detect suspicious users on a real-world experiment?
    
\end{enumerate}
\subsection{Experimental Setup} \label{Subsection: Experimental setup}
\subsubsection{Dataset} \label{Subsubsection: Dataset}
 We use the dataset from Inter Bank, one of the biggest and fastest growing digital banks in Brazil. The dataset is composed of transactions made to/from clients of the bank. Due to privacy reasons, we cannot provide a full detailed description of the dataset's attributes, however, we provide a quantitative description of the data.
 
 The dataset is comprised of a sample of $20$ million users, $200$ transaction types, and $110$ million transactions that span over a month. We split this data into $5$ distinct snapshots, each approximately representing a week. As mentioned before, all snapshots contain the same set of users, however, the number of transactions varies between them. Many of the users in the network are not clients from Inter -- referred to as non-Inter clients --, e.g., clients from other banks or financial institutions. Each Inter client has $16$ tabular attributes, e.g. Personal Income and Address , out of which $13$ are categorical, and all numerical features were standardized. Since Inter does not have tabular information on non-Inter clients, we define their attributes as a vector of ones and also add a dummy variable to all user attributes indicating if the user is an Inter client. Empirical testing showed us that, in the case of \textbf{A2}, it was preferable to keep the dummy variable instead of also making the graph heterogeneous w.r.t.\ its nodes. Classifying such non-Inter clients is beyond the scope of this work, and thus we only keep them during the node embedding generation step (Step~1 in Section \ref{Section: Proposed Framework}) due to their topological value for understanding the overall structure of the network, after which they are no longer considered. 

 The minority class (`suspicious') of the dataset represents if the Inter client was suspected of being involved in money laundering, and thus reported to the authorities by Inter's AML team. The majority class (`non-suspicious') represents Inter clients that: (i) did not trigger any of the rules mentioned in Section \ref{Section:Introduction}, or (ii) triggered a number of rules and, after manual analysis by the AML team, were discarded as a suspect.  The task in question is to detect Inter clients that were suspected to be involved in money laundering activity, and the imbalance ratio between the classes -- the ratio between the number of Inter clients in the minority class and the majority class -- is $2 \cdot 10^{-5}$. The ratio between non-Inter and Inter clients is not disclosed due to privacy reasons.
 
 A natural issue with this type of modeling is that it biases the classes toward the aforementioned fail-prone rule-based system, since there could be -- and in fact there are, as we will show in future sections -- clients that are involved in money laundering and haven't triggered any of the rules, and thus would be mislabeled as `non-suspicious'. A more general approach could then be to create a new class representing unlabeled clients, and then proceed to employ unsupervised methods for labeling. However, there are two main issues with this approach: (i) automatically labeling a great number of real-world clients as suspected of money laundering is a sensitive topic, since some countries -- including Brazil -- demand a manual specialized analysis of the client in order for the authorities to take action, moreover, many of the state-of-the-art models for unsupervised labeling do not have good explainability, making it even harder to argue for their usage with banks in a real-world task, and (ii) the overwhelming majority of clients would be unlabeled, far exceeding the capability of current unsupervised labeling methods.
 
 We highlight that, even though DELATOR does predict probabilities for all Inter clients in the network, it does not perform the automatic labeling of such clients. This step is only made after a manual analysis of the most likely clients by the AML team. More formally, DELATOR is designed to be a \textit{CAAT} (Computer-assisted audit technology), i.e., it is meant to assist the AML team, and not to automate their analysis.
\subsubsection{Baselines} \label{Subsubsection: Baselines}
We now detail the state-of-the-art methods used as baselines. First, we consider GNN baseline methods for extracting node embeddings from graphs, often referred to as embedding generators:
\begin{itemize}
    \item \textbf{DGL-KE}: A state-of-the-art framework developed by \textit{Amazon AWS} for learning representations on knowledge graphs, i.e., heterogeneous graphs with multiple edge and node types.
    \item \textbf{GraphSAGE} (\textbf{SAGE}): A state-of-the-art GNN architecture that generalizes the original model for learning on graphs, by allowing for multiple different aggregator functions on the message passing step.
    \item \textbf{Graph Attention} (\textbf{GAT}): A state-of-the-art GNN architecture that introduces an attention mechanism to the message passing algorithm.
    \item \textbf{Graph Convolution} (\textbf{GCN}): The first proposed GNN architecture based on message passing on graphs.
\end{itemize}

As we will discuss in future sections, the SMOTE step proved to be essential for GNN-based models to perform well, since the large class imbalance makes learning on the available data extremely hard, which also made training models directly on the final supervised classification task ineffective. Thus, we decided to train all of the GNN baseline models according to a similar training procedure as the one described in Section \ref{Section: Proposed Framework}. Specifically, we train the models on an unsupervised link prediction task to obtain node embeddings, then we oversample the minority class on the embedding space for the training set, and last, we train a supervised classifier in the final classification task. We employ approach \textbf{A1} for modeling the data to train the SAGE, GAT, and GCN methods, and approach \textbf{A2} for modeling the data to train DGL-KE. We note that DGL-KE utilizes a modified version of the link prediction task, described in detail in \cite{DGL-KE}.

We also experimented with different architectures of embedding generators for DELATOR for the first step of the framework, i.e., obtaining embeddings according to both the link prediction and edge regression tasks. We denote these variants by including their names as a sub-index, e.g., DELATOR$_{\text{SAGE}}$ denotes the version utilizing GraphSAGE layers on both tasks. We highlight that comparing our framework with the SAGE, GAT and GCN baseline methods will explicit how the edge regression task impacts on the overall performance, since such methods only execute the link prediction task, and comparing DELATOR with DGL-KE will allow us to understand which of the proposed modeling approaches (\textbf{A1} and \textbf{A2}) is best suited for our data.

As a baseline that does not utilize the network's transaction information in its modeling, we consider  \textbf{CatBoost} \cite{catboost}, a state-of-the-art gradient boosting algorithm for supervised learning, that allows for better performance on highly categorical feature spaces. This model learns directly on the client attributes $\mathcal{X}$ in order to obtain the final prediction. Since the raw feature space is mostly categorical, the CatBoost model does not oversample the minority class. There are versions of SMOTE made specifically for dealing with mixed feature spaces, such as SMOTE-NC, however these versions suffer from severe scaling issues, and in our scenario employing them proved to be intractable. Other simpler techniques for dealing with class imbalance, such as utilizing loss functions that increase the penalty for misclassifications in the minority class, also proved to be ineffective and/or inefficient for CatBoost.
\subsubsection{Evaluation Metrics} \label{Subsubsection: Evaluation Metrics}
For evaluating the effectiveness of the proposed framework for detecting suspicious clients in transaction networks, we propose the following metrics:
\begin{itemize}
    \item \textbf{AUC-ROC}: area under the ROC curve (True positive rate vs. False positive rate).
    \item \textbf{AUPR}: area under the PR curve (Precision vs. Recall).
    \item \textbf{F1-Fraud} (\textbf{F1-F}): geometric mean between the precision and recall w.r.t.\ the label that represents suspicious activity.
    \item \textbf{Maximum F1-Fraud} (\textbf{Max. F1-F}): maximum value of the F1-Fraud w.r.t.\ all possible thresholds.
\end{itemize}
\subsubsection{Computing Specifications} \label{Subsubsection: Computing Specifications}
All of the experiments were conducted on the \textit{AWS Sagemaker} platform for cloud-based computing. The experiments for the main framework and baselines using approach \textbf{A1} were conducted on a single \textit{ml.m5.24xlarge} machine, with $96$ Intel® Xeon® vCPUs and $384$ GB of memory. The experiments using DGL-KE were conducted on a single \textit{ml.g4dn.16xlarge} machine, with $64$ Intel® Xeon® vCPUs, $256$ GB of memory and $1$ NVIDIA® V100 Tensor Core GPU with $32$ GB of memory.
\subsubsection{Implementation Details} \label{Subsubsection: Implementation Details}
All GNN models, except for DGL-KE, were trained using the Adam \cite{kingma2014method} optimization algorithm, with a learning rate of $0.01$. These models were trained until the validation loss increased for $5$ consecutive epochs (taking, on average, 40 epochs). A parameter tuning procedure showed that two GNN layers were preferable for all of the aforementioned models, as well as showing that one attention head was the best option for the methods that use the GAT model. We also utilized an embedding size of $60$ per snapshot for DELATOR, with $d_{\text{LP}} = 40$ and $d_{\text{ER}} = 20$ -- and since we have $5$ snapshots, this resulted in a final embedding size of $300$ --, and a negative sample size of $2$ (per positive sample). The remaining baseline GNN models utilized an embedding size of $40$ per snapshot. For all versions of DELATOR, the MLP used in the edge regression step had two hidden layers with dimensions $(120,60)$ and $(60,1)$ respectively, with the ReLU function as a nonlinearity for the first layer. The $\gamma$ parameter of \eqref{eq:er_loss} was set to 3.

DGL-KE provides four different models for representation learning, and we found that TransE\_l$2$ yielded the best results. The model utilizes the edge weights, but by design does not utilize the node features. A parameter tuning procedure showed that higher embedding sizes did not contribute to higher performance, hence an embedding size of $4$ per snapshot was utilized (tested 4, 8, 16, 32, 64), resulting in a final embedding size of $20$. We employed a negative sample size of $128$, and the remaining hyperparameters were initialized with their default values.

Because an extensive grid search over the hyperparameter space revealed little performance improvement, the LightGBM model used in DELATOR's final stage was initialized with its default hyperparameters, as per the official Python implementation. The same procedure was applied to CatBoost's hyperparameter space, and the final version utilized $400$ estimators, a maximum tree depth of $10$ and a learning rate of $1$, using ``log-loss'' as objective, with the remaining hyperparameters being set to their default values.

For the initial embedding generation task (Step~1 in Section \ref{Section: Proposed Framework}), we utilized a 75-10-15 train-test-validation split on the edges of each graph snapshot. For the final supervised task (Step~3 in Section \ref{Section: Proposed Framework}), we utilized an 80-20 train-test split on the Inter clients and then performed a $5$-fold cross validation on the training set. The random number generator seed for this task was also randomized $5$ times in order to obtain $25$ different training configurations to check the statistical significance of our experiments. The numerical values described in tables \ref{tab:overall_results} and \ref{tab:ablation_study} represent the mean results, together with their respective standard deviations, w.r.t\ the proposed metrics.

\subsection{Results} \label{Subsection: Results}
\subsubsection{Quantitative Results} \label{Subsubsection: Quantitative Results}
\begin{table*}[htbp]
\caption{Evaluation results for the final prediction task of detecting suspicious clients.}
\begin{center}
\begin{tabular}{|c|c|c|c|c|}
\hline
\textbf{Method}&\multicolumn{4}{|c|}{\textbf{Evaluation Metrics}} \\
\cline{2-5} 
 & \textbf{\textit{AUC-ROC}} $\uparrow$ & \textbf{\textit{AUPR}} $\uparrow$  & \textbf{\textit{F1-F}} $\uparrow$ & \textbf{\textit{Max. F1-F}} $\uparrow$ \\
\hline
CatBoost& $0.653 \pm 0.075$ & $0.007 \pm 0.015$&  $0 \pm 0$ & $0.009 \pm 0.003$ \\
\hline
DGL-KE& $0.735 \pm 0.028$ & $0 \pm 0$&  $0 \pm 0$ & $0.001 \pm 0$  \\
SAGE& $0.846 \pm 0.039$ & $0.001 \pm 0$&  $0.002 \pm 0$ & $0.019 \pm 0.017$  \\
GAT& $0.743 \pm 0.043$ & $0.001 \pm 0.001$&  $0.001 \pm 0.001$ & $0.025 \pm 0.019$  \\
GCN& $0.876 \pm 0.018$ & $0.002 \pm 0.002$&  $0.002 \pm 0$ & $0.032 \pm 0.025$ \\
\hline
DELATOR$_{\text{SAGE}}$ (Ours)& $\textbf{0.905}\pm 0.027 $ & $0.001 \pm 0$& $\textbf{0.005} \pm 0.001$ & $0.033 \pm 0.010$  \\
DELATOR$_{\text{GCN}}$ \ (Ours)& $0.889 \pm 0.014$ & $\textbf{0.011} \pm 0.019$& $0.003 \pm 0.001$ & $\textbf{0.049} \pm 0.043$  \\
DELATOR$_{\text{GAT}}$ \ (Ours)& $0.882 \pm 0.023$ & $0.001 \pm 0$& $0.003 \pm 0.001$ & $0.018 \pm 0.015$ \\
\hline
\end{tabular}
\label{tab:overall_results}
\end{center}
\end{table*}
To answer the first two questions posed in this section, we compared the results of DELATOR with the proposed baselines for the final task of detecting suspicious clients. 

Table \ref{tab:overall_results} shows the overall evaluation results for the final prediction task. We can observe that, for all metrics, the versions of DELATOR have outperformed the proposed baselines, especially when compared to DGL-KE. The results w.r.t.\ AUC-ROC show that, on average, DELATOR$_{\text{SAGE}}$ provides a more consistent ranking of clients when compared to the other baselines. The AUPR score is also higher for DELATOR$_{\text{GCN}}$, showing that it is easier to select a classification threshold such that the framework's precision dominates its recall. In addition, we note that the F1-Fraud and Maximum F1-Fraud metrics are also higher for both DELATOR$_{\text{SAGE}}$ and DELATOR$_{\text{GCN}}$. This is a strong indicator that the framework has the best performance for correctly classifying clients that are suspected of being involved in money laundering. 

It is worth noting that, although DELATOR$_{\text{GCN}}$ outperforms DELATOR$_{\text{SAGE}}$ w.r.t.\ AUPR and Maximum F1-Fraud, the results for the latter have a considerably smaller standard deviation, i.e., DELATOR's GraphSAGE variant exhibits more consistent performance than the other variants. As for CatBoost, the model also presents a higher standard deviation for the AUPR metric when compared DELATOR$_{\text{SAGE}}$, coupled with lower mean values for the AUC-ROC, F1-Fraud and Maximum F1-Fraud metrics when compared to all other methods, meaning that CatBoost struggles to learn consistently on our dataset.

\begin{figure}[htbp]
\centerline{\includegraphics[height=45mm]{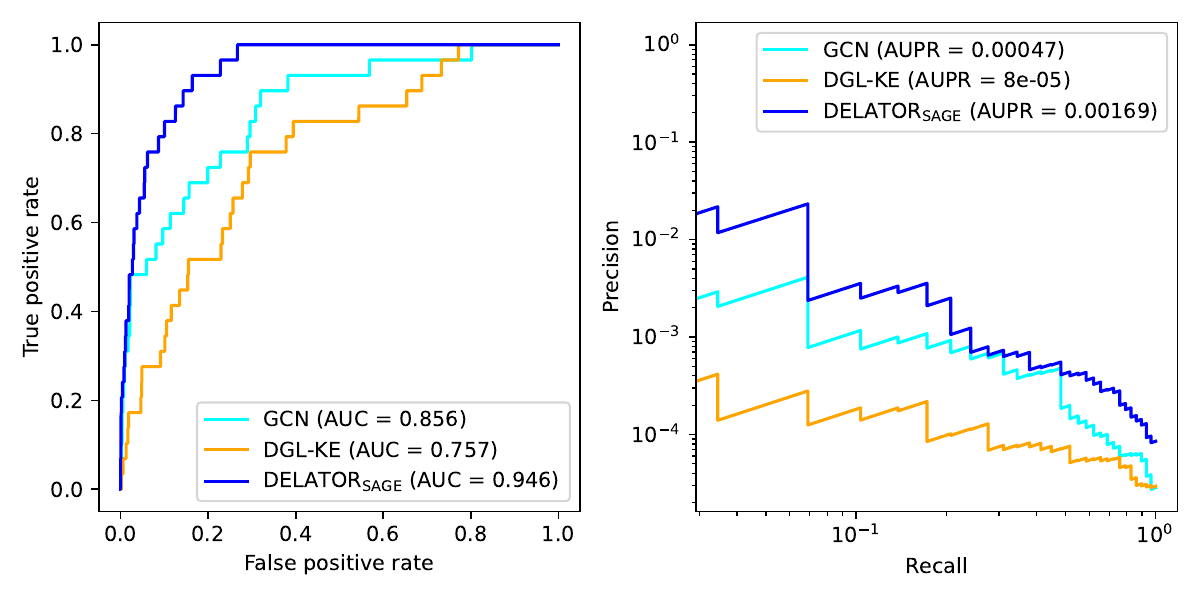}}
\caption{ROC and PR curves for the best versions of DELATOR, GCN and DGL-KE, based on validation sets, for the final task of detecting suspicious clients}
\label{fig:roc_pr_dlt_gcn_ke}
\end{figure}

Fig.~\ref{fig:roc_pr_dlt_gcn_ke} provides the ROC and PR curves for the best versions of DELATOR, GCN, and DGL-KE (based on validation sets). We observe that the ROC curve for DELATOR increases in height at a much faster pace than the baselines, signalling that it is easier to choose a classification threshold for DELATOR such that the true positive rate is high while the false positive rate is reasonably low, i.e., DELATOR is better at detecting suspicious clients. The best version of DELATOR also has a considerably higher AUC-ROC when compared to DGL-KE and GCN. The PR curve for DELATOR shows a much more desirable behavior, as it does not decrease as fast as DGL-KE's or GCN's.

Table \ref{tab:confusion_matrix} shows the confusion matrix for the best versions of DELATOR, GCN, and DGL-KE (based on validation sets). All methods show a similar number of true negatives and false negatives, however, we can see that the number of false positives -- `non-suspicious' clients classified as `suspicious' -- is one order of magnitude greater for DGL-KE when compared to our framework. Moreover, the number of true positives for DELATOR is approximately three times greater than DGL-KE's, and two times greater than GCN's. These results again show how DELATOR outperforms the baselines in detecting suspicious clients.
\begin{table}[htbp]
\caption{Confusion matrix for the best versions of DELATOR, GCN, and DGL-KE, based on validation sets. The best version of DELATOR utilized SAGE layers.}
\begin{center}
\begin{tabular}{|c|c|c|c|c|}
\hline
\textbf{Method}&&&\multicolumn{2}{|c|}{\textbf{Predicted}} \\
\cline{3-5} 
 & &\multicolumn{1}{|c|}{\textbf{Class}}& \textit{non-suspicious} & \textit{suspicious} \\
 \hline 
 DELATOR &\multicolumn{1}{|c|}{}  &\textit{non-suspicious} & $99.806\%$ & $0.191\%$\\
 \cline{3-3}
 & \multicolumn{1}{|c|}{}& \textit{suspicious}&$18\cdot10^{-4}\%$ & $4.7\cdot10^{-4}\%$\\
\cline{1-1}  \cline{3-5}
 GCN &\multicolumn{1}{|c|}{\textbf{Real}} &\textit{non-suspicious} & $99.699\%$ & $0.298\%$\\
 \cline{3-3}
 &\multicolumn{1}{|c|}{} &\textit{suspicious} &$20\cdot10^{-4}\%$ & $2.35\cdot10^{-4}\%$\\
 \cline{1-1}  \cline{3-5}
 DGL-KE &\multicolumn{1}{|c|}{} &\textit{non-suspicious} & $97.977\%$ & $2.021\%$\\
 \cline{3-3}
 &\multicolumn{1}{|c|}{} &\textit{suspicious} &$22\cdot10^{-4}\%$ & $1.6\cdot10^{-4}\%$\\
 \hline
\end{tabular}
\label{tab:confusion_matrix}
\end{center}
\end{table}
\subsubsection{Parameter Sensitivity} \label{Subsubsection: Parameter Sensitivity}
\begin{figure}[htbp]
\small
\centering
\subfloat{\includegraphics[height=1.3in,width=42mm]{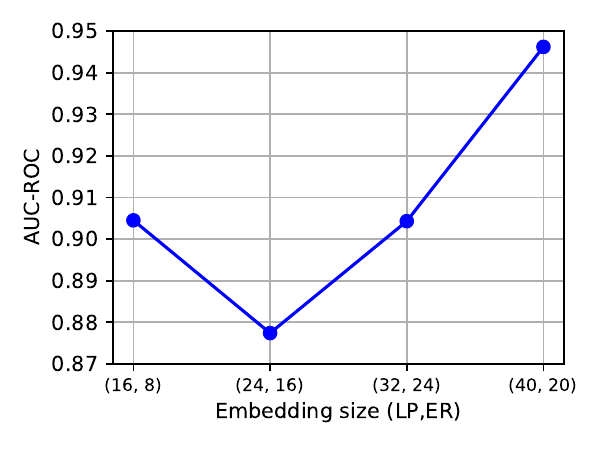} }
\subfloat{\includegraphics[height=1.3in,width=42mm]{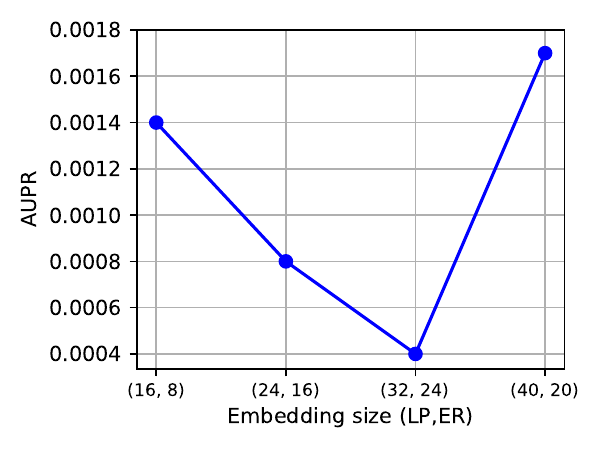} }%
\caption{Impact of embedding size ($d_{\text{LP}}$, $d_{\text{ER}}$) on AUC-ROC and AUPR for the final prediction task of detecting suspicious clients, on a fixed seed, for DELATOR$_{\text{SAGE}}$.} 
\label{fig:embd_size_auc}
\end{figure}

In order to answer the third question posed in this section, we now explore the impact of DELATOR's hyperparameters on the model's performance. More specifically, we analyze the impact of the final embedding size for both GNNs involved in the framework. The GNN trained using the link prediction task has an embedding size of $d_{\text{LP}}$, and the GNN trained using the edge regression task has an embedding size of $d_{\text{ER}}$. Fig.~\ref{fig:embd_size_auc} shows the impact of the embedding size on the AUC-ROC and AUPR metrics. For the AUC-ROC metric, we observe that a larger embedding size significantly benefits the model's ability to classify clients. Nevertheless, these variations are not drastic, thus showing that the model is robust to parameter variations w.r.t.\ the metric. For the AUPR metric, the intermediate values for the embedding size show lower performance, while the extreme values show similarly good results.

\subsubsection{Ablation Study} \label{Subsubsection: Ablation Study}
\begin{table*}[htbp]
\caption{Ablation Study of DELATOR w.r.t.\ the final prediction task of detecting suspicious clients.}
\begin{center}
\begin{tabular}{|c|c|c|c|c|c|}
\hline
\textbf{Embedding Generator}&\textbf{\textit{Uses LP?}}&\textbf{\textit{Uses ER?}}&\textbf{\textit{Uses SMOTE?}}&\multicolumn{2}{|c|}{\textbf{Results}} \\
\cline{5-6}
&&&&\textbf{\textit{AUC-ROC}} $\uparrow$ & \textbf{\textit{AUPR}} $\uparrow$ \\
\hline
& \checkmark&\checkmark&\checkmark&$\textbf{0.905}\pm 0.027 $ & $0.001 \pm 0$ \\
SAGE& \checkmark&\text{\sffamily X}&\checkmark&$0.846 \pm 0.039$ & $0.001 \pm 0$ \\
& \checkmark&\checkmark&\text{\sffamily X}&$0.454 \pm 0.062 $ & $0 \pm 0$ \\
& \checkmark&\text{\sffamily X}&\text{\sffamily X}&$0.290 \pm 0.120$ & $0 \pm 0$ \\
\hline
& \checkmark&\checkmark&\checkmark&$0.889 \pm 0.014$ & $\textbf{0.011} \pm 0.019$ \\
GCN& \checkmark&\text{\sffamily X}&\checkmark&$0.876 \pm 0.018$ & $0.002 \pm 0.002$ \\
& \checkmark&\checkmark&\text{\sffamily X}&$0.471 \pm 0.062$ & $0 \pm 0$ \\
& \checkmark&\text{\sffamily X}&\text{\sffamily X}&$0.497 \pm 0.001$ & $0 \pm 0$ \\
\hline
& \checkmark&\checkmark&\checkmark&$0.882 \pm 0.023$ & $0.001 \pm 0$ \\
GAT& \checkmark&\text{\sffamily X}&\checkmark&$0.743 \pm 0.043$ & $0.001 \pm 0.001$ \\
& \checkmark&\checkmark&\text{\sffamily X}&$0.502 \pm 0.045$ & $0 \pm 0$ \\
& \checkmark&\text{\sffamily X}&\text{\sffamily X}&$0.440 \pm 0.036$ & $0.003 \pm 0$ \\
\hline
\end{tabular}
\label{tab:ablation_study}
\end{center}
\end{table*}
In order to answer the fourth question posed in this section, we perform an ablation study of DELATOR's components detailed in Table~\ref{tab:ablation_study}. We note that, since all baseline GNN models were trained following the steps described in Section \ref{Section: Proposed Framework}, the results in Table~\ref{tab:overall_results} w.r.t.\ GAT, GCN, and SAGE models when compared to the versions of DELATOR already provide the necessary information for underlining the importance of the edge regression task for the final prediction.

Table \ref{tab:ablation_study} shows that the SMOTE step is crucial for the performance of the framework in the final task, since all versions of the framework trained without this step resulted in models with AUC-ROC score close to or below $0.5$, i.e., the models were at most as good as random guessing. We can also see that, for all models, the edge regression task does provide a considerable improvement to the final classification scores when compared to the baseline models that were only trained on the link prediction task. Again, this shows evidence of the superiority of our approach when compared to the considered baselines. 

\begin{figure}[htbp]
\centerline{\includegraphics[height=45mm]{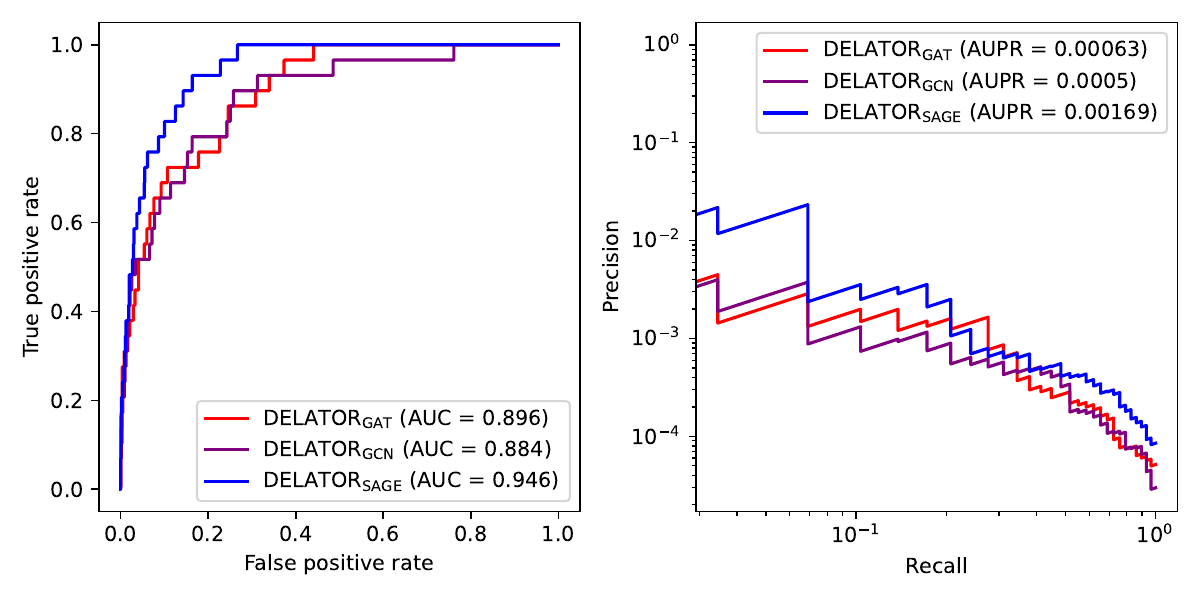}}
\caption{ROC and PR curves for different versions of DELATOR, on a fixed seed, for the final task of detecting suspicious clients.}
\label{fig:roc_pr_dlt_only}
\end{figure}

Fig.~\ref{fig:roc_pr_dlt_only} shows the ROC and PR curves for each variant of DELATOR. We conclude that DELATOR's GraphSAGE variant exhibits the best performance, since both curves display the best behavior for choosing a good classification threshold. 

\subsubsection{Real-World Experiment} \label{Subsubsection: Real World Experiment}
In order to answer the final question posed in this section, we also conducted a real-world experiment with Inter to verify the performance of DELATOR. As previously discussed, most Brazilian banks implement a flawed rule-based system for detecting clients suspected of being involved in money laundering. One of the most significant issues with this system is the fact that the rules themselves are not defined according to a data-driven procedure, meaning that many criminals who do not fit the human-defined rules are not considered for analysis. Since most banks in Brazil work in a similar manner, they lack a good way of ranking which clients labelled as `non-suspicious' should be analyzed in order to detect possible failures in the rules. 

This is where DELATOR can provide significant help for the AML team. Due to the framework's low training and prediction time, it produce a list of most likely suspicious clients effectively. In this context, we perform an experiment consisting of ranking all clients that did not trigger any alarms during a given time interval. We consider the top $50$ clients with highest probability of being classified as `suspicious' according to DELATOR. These clients were manually analyzed by the AML team, which resulted in the discovery of $7$ new clients involved in suspicious activity, that were then reported to the authorities. This experiment shows that DELATOR can create a new effective and efficient way of detecting suspicious clients that would be otherwise mislabelled as `non-suspicious' by the rule-based system, as well as providing a way of detecting possible rule failures which can then be updated by the analysts, thus vastly enhancing the capabilities of the AML team. 

\section{Privacy and Ethical Considerations}
In this section, we provide a few comments regarding the ethical details of our work. We note that the proposed framework does not violate ethical aspects of the relationship between Inter and its clients, since Brazilian banks have the legal obligation to investigate financial crimes. This work also does not violate the privacy of clients, since all of the data was provided in an anonymized form, according to the recommendations of Inter's data privacy team. Moreover, the entire data processing procedure was made on the Amazon AWS Sagemaker platform from computers that were provided by Inter and that implement a myriad of security techniques, e.g.\ using an exclusive VPN service.

\section{Conclusion and Future Work}
In this work, we introduce a novel framework for detecting money laundering activity in large transaction graphs. By leveraging different information available in the network, we are able to construct a multi-task learning procedure that generates time-aware node representations on euclidean space, which then allows us to perform a supervised learning procedure in order to predict the probability of a given client being involved in money laundering. This procedure is robust to highly imbalanced target classes, allowing for an effective method for learning on dynamic networks that suffer from large class imbalance. We then perform a series of experiments to evaluate the performance of our model, and experimentally demonstrate that DELATOR outperforms all considered baselines w.r.t.\ the evaluation metrics. We also perfom a real-world experiment with the help of Inter's AML team to evaluate DELATOR's performance in a practical scenario, and show that the model was able to detect multiple clients involved in suspicious activity. Overall, we have shown that DELATOR provides an efficient and effective way of detecting money laundering activity and that the framework can provide significant help to banks and financial institutions.

In the future, we plan to test DELATOR in other datasets related to money laundering detection, such as the datasets from the AMLSim project \cite{AMLSim}. We also intend to test the generalization capabilities of the framework in other tasks related to financial activities, such as financial fraud detection. 
\bibliographystyle{IEEEtran}
\bibliography{main}
\end{document}